\documentclass{llncs}
\usepackage{isolatin1}
\usepackage{array}
\usepackage{moreverb}
\usepackage{tabularx}
\usepackage{graphicx}
\usepackage{amsmath}
\usepackage{hyperref}
\def\RR{{\rm I\hspace{-0.50ex}R}}

\graphicspath{{figures/}}

\def\RR{{\rm I\hspace{-0.50ex}R}}
\def\esp{{\hspace*{.5in}}}

\def\4d{{\em 4D-Miner}}

\title{Functional Brain Imaging with Multi-Objective Multi-Modal Evolutionary Optimization}
\titlerunning{Functional Brain Imaging with MoMOO EC}

\author{Vojtech Krmicek\inst{1,2} \and Mich\`ele Sebag\inst{2}}
\authorrunning{Krmicek and Sebag}
\tocauthor{Vojtech Krmicek (Masaryk University) and Mich\`ele Sebag (CNRS)}
\institute{
Department of Computer Science\\
Masaryk University, CZ-602 00 Brno\\
\and
IA-TAO, CNRS $-$ INRIA $-$ LRI\\
Universit\'e Paris Sud, FR-91405 Orsay\\
\email{\{krmicek,sebag\}@lri.fr}\\[1.2em]
}

\begin{document}

\maketitle

\begin{abstract}
Functional brain imaging is a source of spatio-temporal data mining
problems. A new framework hybridizing 
multi-objective  and multi-modal optimization is proposed  
to formalize these data mining problems, and addressed through 
Evolutionary Computation (EC). 

The merits of EC for spatio-temporal data mining
are demonstrated as the approach facilitates the modelling of the experts'
requirements, and flexibly accommodates their changing goals.
\end{abstract}

\section{Introduction}
Functional brain imaging aims at understanding the 
mechanisms of cognitive processes through non-invasive technologies
such as magnetoencephalography (MEG). These technologies measure 
the surface activity of the brain with a good spatial and temporal resolution 
\cite{Hamalainen,Baillet}, generating massive amounts of data. 

Finding ``interesting'' patterns in these data, e.g. assemblies 
of active neuronal cells, can be viewed as a Machine Learning or a 
Data Mining problem. However, contrasting with  ML or DM 
applications \cite{Hastie}, the appropriate search criteria are
not formally defined up to now; in practice the detection of active cell assemblies is 
manually done.

Resuming an earlier work \cite{IJCAI05}, 
this paper formalizes functional brain imaging as a 
multi-objective multi-modal optimization (MoMOO) problem, and describes the
evolutionary algorithm called \4d\ devised to tackle this problem. 
In this paper, the approach is extended to the search of discriminant patterns;
additional criteria are devised and accommodated in order to find 
patterns specifically related to particular cognitive activities.

The paper is organized as follows. Section \ref{position} introduces the 
background and notations; it describes the targeted spatio-temporal
patterns (STP) and formalizes the MoMOO framework proposed.
Section \ref{alg} describes the \4d\ 
algorithm designed for finding STPs, hybridizing multi-objective \cite{Debbook}
and multi-modal \cite{Multi-modal} heuristics, and it reports 
on its experimental validation.
Section \ref{disc} presents 
the extension of \4d\ to a new goal, the search for discriminant STPs.
Section \ref{state} discusses the opportunities offered by Evolutionary 
Data Mining, and the paper concludes with perspectives for
further research.

\section{Background and Notations}\label{position}
\begin{figure}[htb]
    \centerline{\includegraphics[width=3in,height=1.5in]{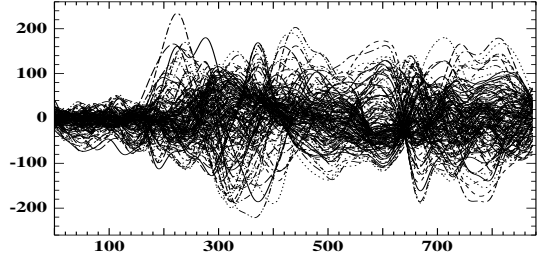}}
    \caption{Magneto-Encephalography Data (N = 151, T = 875)}
\label{Fig-data}
    \end{figure}

This section introduces the notations and criteria 
for Data Mining in functional brain imaging, assuming the reader's familiarity
with multi-objective optimization \cite{Debbook}.
Let $N$ be the number of sensors and let $T$ denote the 
number of time steps. The $i$-th sensor is characterized by
its position $M_i$ on the skull ($M_i = (x_i,y_i,z_i) \in \RR^3$) and 
its activity $C_i(t), 1 \leq t \leq T$ along the 
experiment. Fig. \ref{Fig-data} depicts a set of 
activity curves.

A spatio-temporal pattern noted $X = (I,i,w,r)$ 
is characterized from  its temporal interval $I$ 
($I = [t_1,t_2] \subset [1,T]$) and a spatial region  ${\cal B}(i,w, r)$. 
For the sake of convenience, spatial regions are restricted to 
axis-parallel ellipsoids centered on some sensor; region
 ${\cal B}(i,w, r)$ is the ellipsoid centered on the $i$-th sensor, which  includes
 all sensors $j$ such that $d_w(M_i,M_j)$ is less than radius $r>0$, with 
\[ d_w(M_i,M_j)^2 = w_1 (x_i - x_j)^2 + w_2 (y_i - y_j)^2 + w_3 (z_i - z_j)^2 
\esp w_1,w_2,w_3 > 0 \] 

This paper focuses on the detection of assemblies of active neuronal cells, informally
viewed as large spatio-temporal regions with correlated
sensor activities.  
Formally, let $I = [t_1,t_2]$ be a time interval, and let $\bar C_{i}^I$
denote the average activity of the $i$-th sensor over $I$. 
The $I$-alignment $\sigma_I(i,j)$ of sensors $i$ and $j$ over $I$ 
is defined as:
\[
 \sigma_I(i,j)~=  \frac{\sum_{t=t_1}^{t_2} C_i(t) . C_j(t)}{\sqrt{\sum_{t=t_1}^{t_2}C_i(t)^2} ~\times~\sqrt{\sum_{t=t_1}^{t_2}C_{j}(t)^2 }} \times \left(1 - \frac{|\bar C_i^I - \bar C_j^I|}{|\bar C_i^I|}\right), 
\]
To every spatio-temporal pattern $X = (I,i,w,r)$, are thus associated 
i) its duration or length $\ell(X)$ ($= t_2-t_1$); ii) its area 
 $a(X)$ (the number of sensors in ${\cal B}(i,w, r)$); and iii) its 
alignment $\sigma(X)$, defined as the average 
of $\sigma_I(i,j)$ for $j$ ranging in ${\cal B}(i,w,r)$. 
An {\em interesting} candidate pattern is one with large length, area and alignment.

Naturally, the sensor alignment tends to decrease as a longer time interval or 
a larger spatial region are considered, everything else being equal;
conversely, the alignment increases when the duration or the area decrease. 
It thus comes to characterize the STP detection problem as a multi-objective 
optimization problem  (MOO) 
\cite{Debbook}, searching for large spatio-temporal regions $X$ with 
correlated sensor activities, i.e. patterns $X$ simultaneously 
maximizing criteria $\ell(X),~ a(X)$ and $\sigma(X)$. The best compromises 
among these criteria, referred to as Pareto front, are the solutions of the 
problem. \smallskip\\
\noindent{\bf Definition 1. (Pareto-domination)}\\
{\em Let $c_1,\ldots,c_K$ denote $K$ 
criteria to be simultaneously maximized on $\Omega$. 
$X$ is said to Pareto-dominate $X'$ if
$X$ improves on $X'$ with respect to all criteria, and the improvement is strict
for at least one criterion.
The Pareto front includes all solutions which are 
not Pareto-dominated.
}\smallskip\\
However, the MOO setting fails to capture the true target patterns: The 
Pareto front defined from the above three criteria 
could be characterized and it does include a number of patterns; but all of these 
actually represent the same spatio-temporal region up to some slight variations 
of the time interval and the spatial region. This was found unsatisfactory as 
neuroscientists are actually interested in {\em all} active areas of the brain;
 $X$ might be worth even though its alignment, duration and area 
are lower than that of  $X'$, provided that $X$ and $X'$ are 
situated in different regions of the brain.  

The above remark leads to extend  multi-objective optimization goal 
in the spirit of multi-modal optimization \cite{Multi-modal}. Formally, 
a new optimization framework is defined,  
referred to as {\em multi-modal multi-objective optimization} (MoMOO). 
MoMOO uses a relaxed inclusion relationship, 
noted $p$-inclusion, to relax the Pareto domination relation.\smallskip\\
\noindent{\bf Definition 2. (p-inclusion)}\\
{\em Let $A$ and $B$ be two subsets of a measurable set $\Omega$, and 
let $p$ be a positive real number ($p \in [0,1]$). $A$ is $p$-included in 
$B$ iff  $| A \bigcap B| > p \times |A| $, where $|A|$ 
denotes the measure of set $A$.
}\smallskip\\
\noindent{\bf Definition 3. (multi-modal Pareto domination)}\\
{\em Let $X$ and $Y$ denote two spatio-temporal patterns
with respective supports $Sup(X)$ and $Sup(Y)$ ($Sup(X), Sup(Y) \subset \RR^d$). 
$X$ p-mo-Pareto dominates $Y$ iff the 
 support of $Y$
is $p$-included in that of $X$, and $X$ Pareto-dominates $Y$.
}\smallskip\\
Finally, the interesting STPs are all spatio-temporal patterns which are 
not $p$-mo-Pareto dominated. 

It must be emphasized that MoMOO differs from MOO with diversity enforcing heuristics
(see e.g.,  \cite{Corne,Laumans}): diversity-based heuristics in MOO aim
at a better sampling of the Pareto front defined from fixed objectives; 
MoMOO is interested in a new Pareto front, including diversity as a new objective.

\section{4D-Miner}\label{alg}
This section  describes the \4d\ algorithm designed for the 
detection of stable spatio-temporal patterns, and reports on its experimental
validation.

\subsection{Overview of 4D-Miner} \label{init}
Following \cite{DaidaGECCO99}, special care is devoted to the initialization step.
In order to both favor the generation of relevant STPs and exclude the extremities of 
the Pareto front (patterns with insufficient alignment, or insignificant 
spatial or temporal amplitudes), 
every initial pattern $X =(i,w,I,r)$ is generated after a constrained
sampling mechanism:
\begin{itemize}
\item Center $i$ is uniformly drawn in $[1,N]$;
\item Vector $w$ is set to $(1,1,1)$ ($d_w$ is initialized to 
the Euclidean distance);
\item Interval $I = [t_1,t_2]$ is such that $t_1$ is drawn with uniform 
distribution in $[1,T]$; the length $t_2-t_1$ of $I_j$ is drawn 
according to a Gaussian distribution 
${\cal N}(min_\ell, min_\ell/10)$, where $min_\ell$ is a user-supplied 
length parameter.
\item Radius $r$ is deterministically computed from a 
 user-supplied threshold $min_\sigma$, corresponding to the 
minimal $I$-alignment desired. 
\[ r = min_k \{ d_w(i,k)~s.t.~ \sigma_I(i,k) > min_\sigma) \} \]
\item Last, the  spatial amplitude $a(X)$ of individual $X$ is required to 
be more than a user-supplied threshold $min_a$; otherwise, the individual 
is non admissible and it does not undergo mutation or crossover.
\end{itemize}
The user-supplied $min_\ell$, $min_\sigma$ and $min_a$ thus 
govern the proportion of admissible individuals in the initial population.
The computational complexity of the initialization phase is  
${\cal O}(P \times N \times min_\ell)$, where $P$ is the population size, 
$N$ is the number of measure points and $min_\ell$ is the average length 
of the intervals. 

The variation operators go as follows. From parent $X=(i,w,I,r)$, mutation generates an 
offspring by one among the following operators:
i) replacing center $i$ with another sensor in ${\cal B}(i,w,r)$;
ii) mutating $w$ and $r$ using self-adaptive Gaussian mutation; 
iii) incrementing or decrementing the bounds of interval $I$;
iv) generating a brand new individual (using the initialization operator).\\
The crossover operator is subjected to restricted mating (only sufficiently close 
patterns are allowed to mate); it proceeds by
i) swapping the centers or ii) the ellipsoid coordinates of the two individuals, 
or iii) merging the time intervals.

A steady state evolutionary scheme is considered. In each step, a single admissible
parent individual is selected and it generates an offspring via mutation or crossover;
the parent is selected using a Pareto archive-based selection \cite{Debbook}, 
where the size of the Pareto archive is 10 times the population size. The
offspring either replaces a non-admissible individual, or an individual selected 
after inverse Pareto archive-based selection.

\subsection{Experimental results}\label{setting} \label{expe}
This subsection reports on the experiments  
 done using \4d on real-world datasets\footnote{
Due to space limitations, the reader is referred to  \cite{IJCAI05}
for an extensive validation of \4d. The retrieval performances and 
scalability were assessed on artificial datasets,  varying the 
number $T$ of time steps and the number $N$ of sensors up to 8,000 and 4,000
respectively; the corresponding computational runtime (over a 456Mo dataset) 
is 5 minutes on PC-Pentium IV.}, 
collected from subjects observing a moving ball. Each dataset involves 151 
measure points and the number of time steps (milliseconds) is 875. 
As can be noted from Fig. \ref{Fig-data}, the range 
of  activities widely varies along time.
The runtime on the available data is less than 20 
seconds on PC Pentium 2.4 GHz.

The parameters used in the experiments are as follows. The 
population size is $P=200$; the stop criterion is based on the number of 
fitness evaluations per run, limited to 40,000. 
A few preliminary runs were used 
to adjust the operator rates; the mutation and crossover rates 
are respectively set to .7 and .3.
For computational efficiency, the $p$-inclusion is computed as: $X$ is
$p$-included in $Y$ if the center $i$ of $X$ belongs to the spatial 
support of $Y$, and there is an overlap between their time intervals.
\4d\ is written in C$^{++}$. 

    \begin{figure}[htb]
\centerline{\begin{tabular}{cc}
{\includegraphics[width=2.5in,height=1.4in]{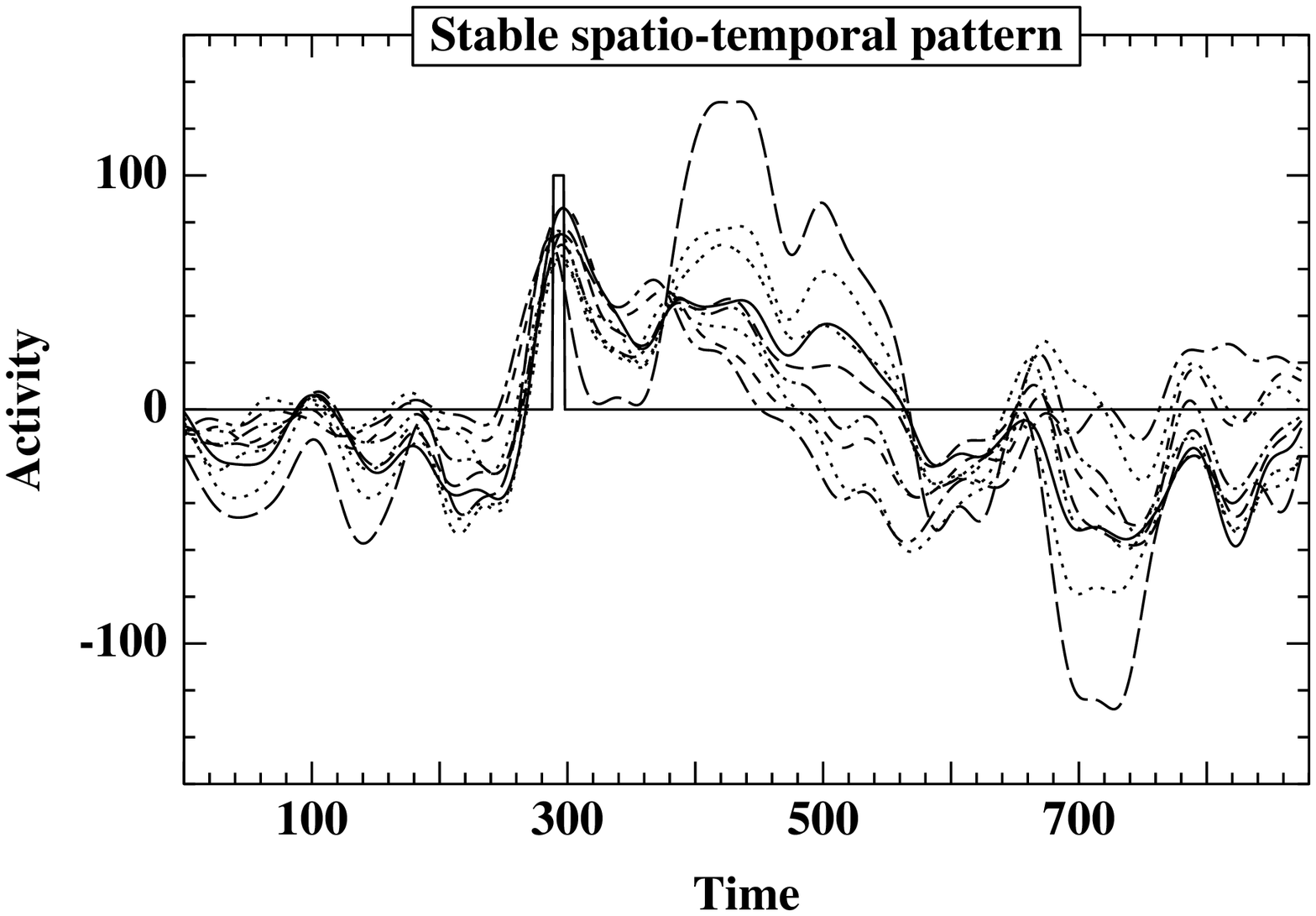}}
&
{\includegraphics[width=2.5in,height=1.4in]{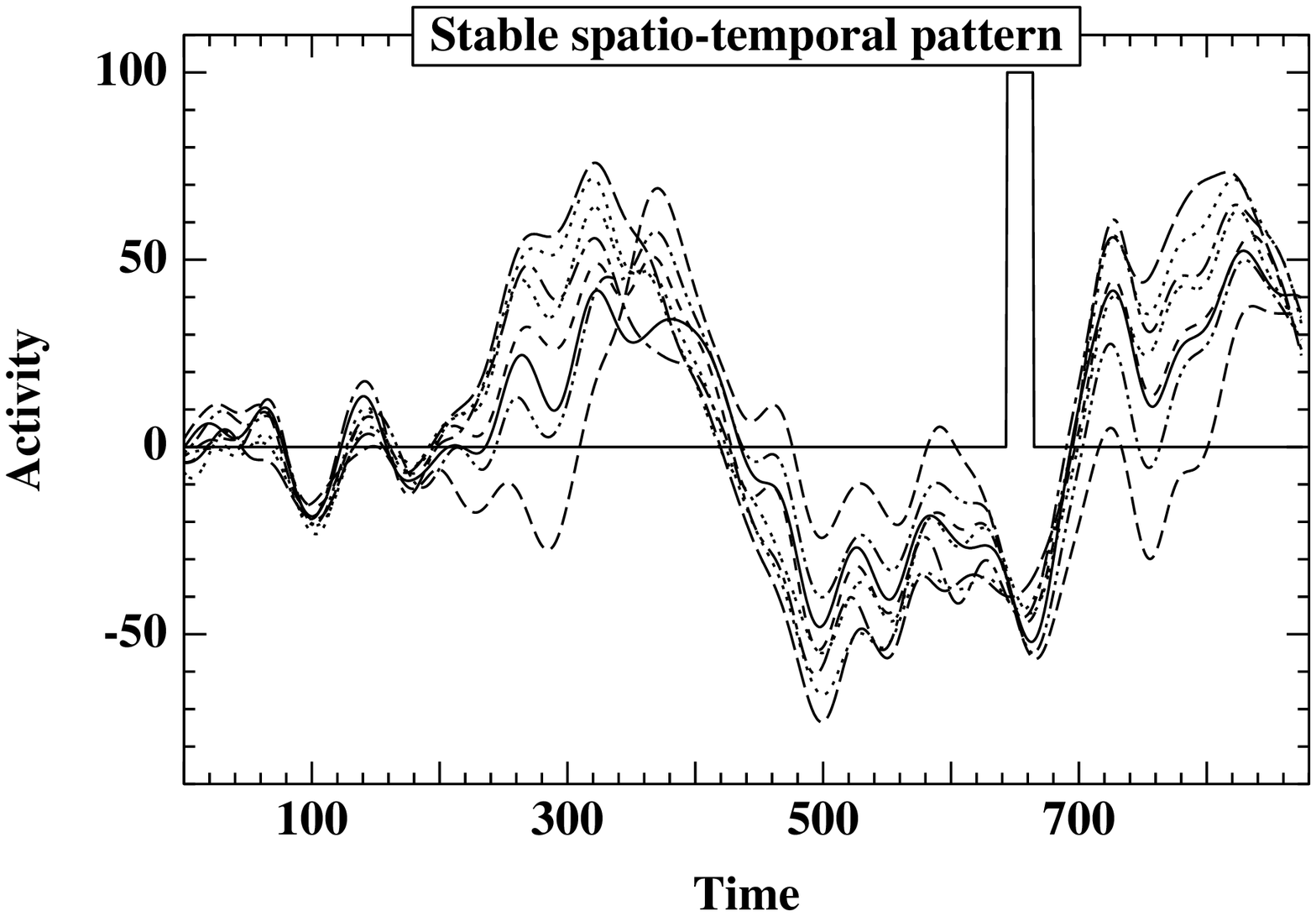}}\\
(a): $\ell(X) = 8, ~a(X)=8, ~\sigma(X) =  .29$ & 
(b): $\ell(X) = 20, ~a(X)=9, ~\sigma(X) =  .396$ \\ 
\end{tabular}}
    \caption{Two stable spatio-temporal patterns ($N = 151,~T=875$)}\label{STP1}
    \end{figure}

Typical STPs found in the real datasets are shown in 
Fig. \ref{STP1}.(a) and (b), displaying all activity curves belonging
to the STP plus the time-window of the pattern. 
Both patterns are considered relevant by the expert; note that 
the STP on the right is Pareto dominated by the one of the left.

All experiments confirm the importance of the user-defined thresholds ($min_\ell$, $min_a, ~min_\sigma$), defining the minimum requirements on solution individuals.
Raising the thresholds beyond certain values leads to poor final results as
the optimization problem becomes over constrained; lowering the thresholds leads to a 
crowded Pareto archive, increasing the computational time and adversely affecting the
quality of the final solutions.
Indeed, the coarse tuning of the parameters can be achieved based on 
the desired proportion of admissible individuals in the initial population. 
However, the fine-tuning
of the parameters could not be automatized up to now, and it 
still requires running \4d\ a few times.
For this reason, the control of the computational 
cost (through the population size and number of generations) is of 
utmost importance.

\section{Extension to Discriminant STP\lowercase{s}} \label{disc}
After some active brain areas have been identified, the next task in the functional brain imaging
agenda is to relate these areas to specific
cognitive processes, using contrasted experimental 
settings. In this section,  the 
{\em catch} versus {\em no-catch} experiment is 
considered;  the subject sees a ball, which s/he
must respectively catch ({\em catch} setting) or 
let go ({\em no-catch} setting). Cell assemblies that 
are found active in the {\em catch} setting and 
inactive in the {\em no-catch} one, are conjectured 
to relate to motor skills. 

More generally, the mining task  becomes to find STPs 
that behave differently in a pair of 
(positive, negative) settings, 
referred to as discriminant STPs. 
The notations are modified as follows. To 
the $i$-th sensor are attached its 
activities in the positive and negative settings,  respectively
noted  $C_i^+(t)$ and $C_i^{-}(t)$; its positions 
are similarly noted $M_i^+$ and $M_i^{-}$.

The fact that the sensor position differs depending on 
the setting entails that the genotype
of the sought patterns must be redesigned. An 
alternative would have been to 
specify the 3D coordinates of a pattern  instead
of centering the pattern on a sensor position. 
However, the spatial region of a pattern actually
corresponds to a set of sensors; in other words 
it is a discrete entity. The use of a 3D (continuous)
spatial genotype would thus require to redesign the 
spatial mutation operator, in order to ensure 
effective 
mutations. However, calibrating the
continuous mutation operator and finding the right 
trade-off between ineffective and disruptive 
modifications of the pattern position proved to be 
trickier than extending the genotype.

Formally, the STP genotype noted $X(i,j,I,w,r)$
now refers to a pair of sensors $i,j$, which are closest to each other across both settings\footnote{
With $j = arg~min \{ d_w(M^+_i,M^-_k), k = 1.. N \}; ~i = arg~min \{ d_w(M^+_k,M^-_j), k = 1.. N \}$.}.
The STP is assessed from:
\begin{itemize}
\item its spatial amplitude $a^+(X)$ (resp. $a^-(X)$) 
defined as the size of ${\cal B}^+(i,w,r)$, including 
all sensors $k$ such that $d_w(M^+_i,M^+_k) < r$ 
(resp. ${\cal B}^-(j,w,r)$, including 
all sensors $k$ such that $d_w(M^-_j,M^-_k) < r$)).  
\item its spatio-temporal alignment  $\sigma^+(X)$
(respectively  $\sigma^-(X)$), defined as the 
activity alignment of the sensors in ${\cal B}^+(i,w,r)$ (resp. in  ${\cal B}^-(j,w,r)$), over time interval $I$.
\end{itemize}

The next step regards the formalization of the goal.
Although neuroscientists have a clear
idea of what a discriminant STP should look 
like, turning this idea into a set of operational
requirements is by no way easy.
Several formalizations were thus considered, 
modelling the search criteria in terms of 
new objectives 
(e.g. maximizing the difference between $\sigma^+$ and
$\sigma^-$) or in terms of constraints 
($|\sigma^+(X) - \sigma^-(X)| > min_{d\sigma}$).
The extension of the \4d\ system to accommodate the 
new objectives and constraints was straightforward. 

The visual inspection of the results found along the 
various modellings led the neuroscientists
to introduce a new feature noted $d(X)$, the 
difference of the average activity in 
${\cal B}^+(i,w,r)$ and ${\cal B}^-(j,w,r)$ over the 
time interval $I$. Finally, the search 
goal was modelled as an additional constraint 
on the STPs,
expressed as $|d(X)| > min_d$ where $min_d$ is a 
user-supplied threshold.
 
Also, it was deemed neurophysiologically unlikely 
that a functional difference could
 occur in the early brain signals; only differences occurring after 
the motor program was completed by the subject, 
i.e. 200ms after the
 beginning of the experiment, are considered 
to be relevant. This requirement was expressed in 
a straightforward way, through a new constraint on 
admissible STPs, and directly at the initialization level (e.g., drawing $t_1$ uniformly in $[200,T]$, section 
\ref{init}).

\begin{figure}
\centerline{\begin{tabular}{cc}
{\includegraphics[width=2.5in,height=1.6in]{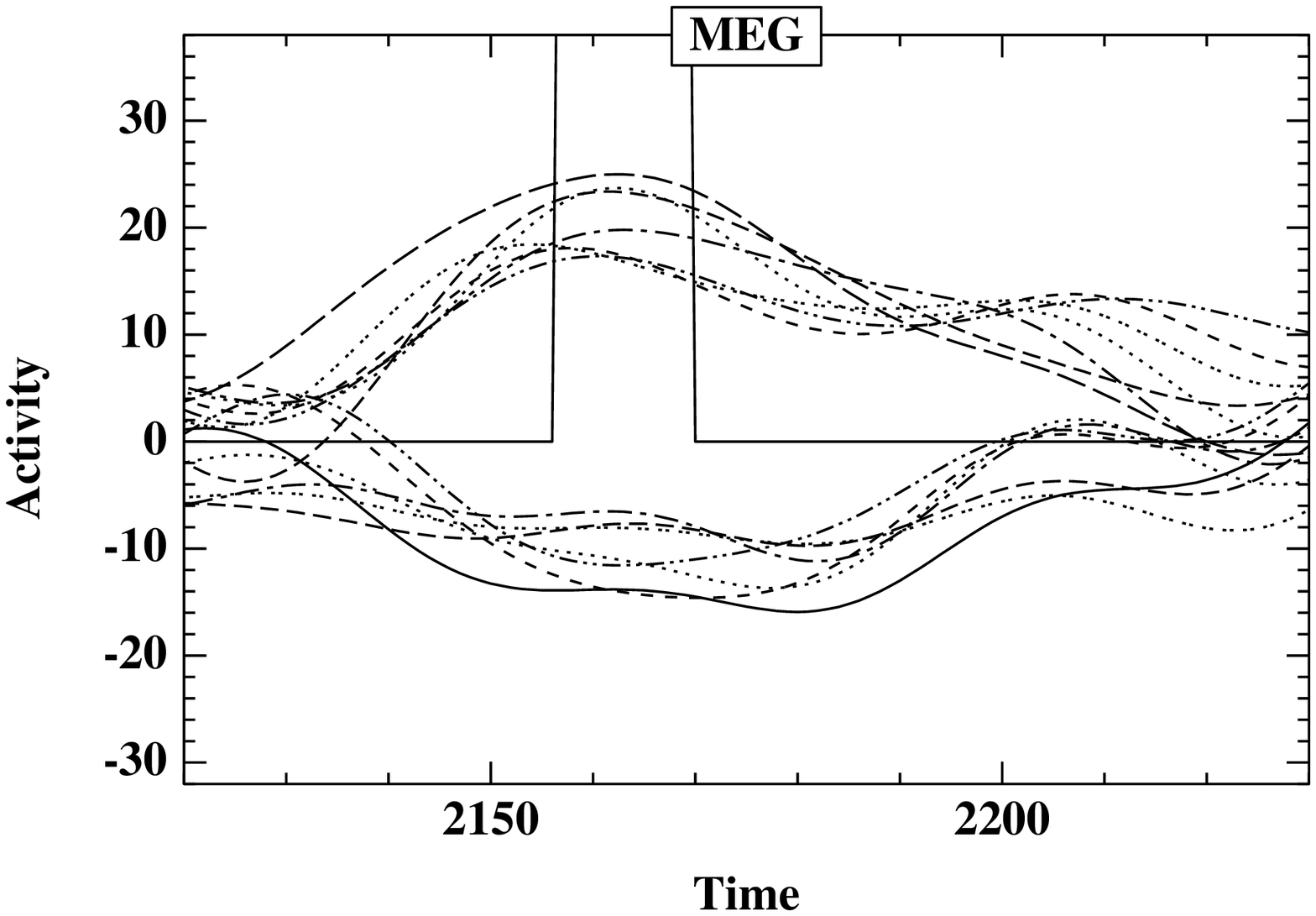}} & 
{\includegraphics[width=2.5in,height=1.6in]{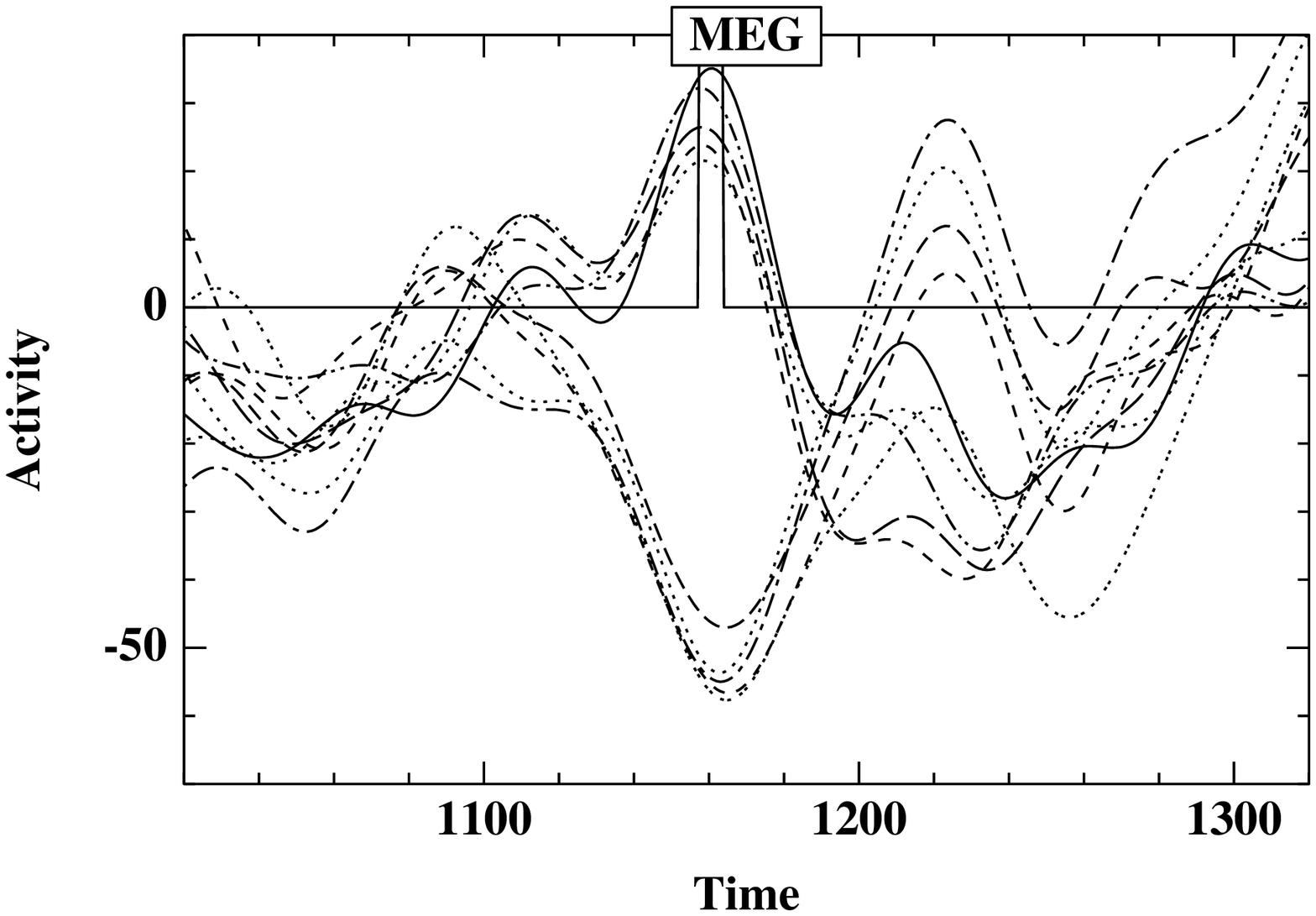}} \\
(a) & (b)\\
\end{tabular}}
    \caption{Two discriminant stable spatio-temporal patterns. Sensors display a positive 
(resp. negative) activity in the catch (resp. nocatch) case. ($N=151$, $T = 2500$)} \label{fdisc}
    \end{figure}

Figs. \ref{fdisc}.(a) and  \ref{fdisc}.(b) show 
two discriminant patterns, that were found 
to be satisfactory by the neuroscientists. Indeed, 
this assessment of the results pertains to the 
field of data mining more than discriminant learning.  
It is worth mentioning that the little amount of 
data available in this study, plus the known 
variability of brain activity in the general case
(between different persons and for a same person
at different moments, see e.g. \cite{BCIPhD}), 
does not permit to assess 
discriminant patterns (e.g. by splitting the data into
training and test datasets, and evaluating the 
patterns extracted from the training set onto the
test set).

Overall, the extension of \4d\ to the search of discriminant STPs 
required i) a small modification of the genotype and ii) the modelling of 
two additional constraints. An additional parameter
was introduced, the minimum difference $min_d$ on the activity level,
which was tuned by a few preliminary runs. Same parameters as in 
section \ref{expe} were used; the computational cost is less than 25 seconds
 on PC-Pentium.

\section{State of the art and discussion}\label{state}

The presented approach is concerned with finding specific 
patterns in databases describing spatial objects along time.

Many approaches have been developed in signal processing
 and computer science to address such a goal, 
ranging from Fourier Transforms to Independent Component 
Analysis \cite{ICA} and mixtures of models \cite{Padraic}.
These approaches aim at particular pattern properties (e.g. independence, generativity) and/or focus on particular data 
characteristics (e.g. periodicity).

Functional brain imaging however does not fall within the range of such 
wide spectrum methods, for two reasons. Firstly, the sought spatio-temporal
patterns are not periodic, and not independent. Secondly, and most importantly,
it appears useless
to build a general model of the spatio-temporal activity, while the ``interesting'' 
activity actually corresponds to a minuscule fragment of the total activity $-$ the
proverbial needle in the haystack.

In the field of spatio-temporal data mining (see 
\cite{Spatial,TEMPORAL} for  comprehensive surveys), 
typical applications such as remote sensing, environmental studies, 
or medical imaging, involve complete algorithms, 
achieving an exhaustive search or building a global model. The stress is 
put on the scalability of the approach.

Spatio-temporal machine learning mostly focuses on clustering, 
outlier detection, denoising, and trend analysis. For instance, 
\cite{Padraic} used EM algorithms for non-parametric 
characterization of functional data (e.g. cyclone trajectories), 
with special care regarding the invariance of the models 
with respect to temporal translations. The main limitation
of such non-parametric models, including Markov Random Fields,
is their computational complexity; therefore the use of 
randomized algorithms is attracting an increasing  for
 sidestepped by using randomized
search for model estimates.

Many developments are targeted at efficient access primitives
and/or complex data structures (see, e.g., \cite{Yu}); another
line of research is based on visual and interactive data mining
(see, e.g., \cite{Keim}), exploiting the 
unrivaled capacities of human eyes for spotting regularities in 
2D-data.

More generally, the presented approach can be discussed with respect to the 
{\em generative} versus {\em discriminative} dilemma in Machine Learning. 
Although the learning goal is most often one of discrimination, 
generative models often outperform discriminative approaches, particularly
when considering low-level information, e.g. signals, images or videos (see e.g. \cite{Bengio}). The higher 
efficiency of generative models is frequently explained as they enable the 
modelling and exploitation of domain knowledge in a powerful 
and convenient way,  ultimately 
reducing the search space by several orders of magnitude. 

In summary, generative ML extracts faithful models of the phenomenon at hand, 
taking advantage of whatever prior knowledge is available; these models 
can be used for discriminative purposes, though discrimination is not among the 
primary goals of generative ML. In opposition, discriminative ML focuses on 
the most 
discriminant hypotheses in the whole search space; it does not 
consider the relevance 
of a hypothesis with respect to the background knowledge {\em per se}. 

To some extent, the presented approach combines generative and discriminative ML.
\4d\ was primarily devised with the extraction of interesting patterns in mind.
The core task was to model the prior knowledge through relevance criteria, 
combining optimization objectives (describing the expert's preferences)
and constraints (describing what is {\em not} interesting). The extraction of 
discriminant patterns from the relevant ones was relatively straightforward, 
based on the use of additional objectives and constraints. This suggests that extracting discriminant
patterns from relevant ones is much easier than searching discriminant patterns, 
and thereafter sorting them out to find the relevant ones.

\section{Conclusion and Perspectives}

This paper has proposed a stochastic approach for mining 
stable spatio-temporal patterns. Indeed, a very simple alternative 
would be to discretize the spatio-temporal domain and 
compute the correlation of the signals in each cell of 
the discretization grid. 
However, it is believed that the proposed approach presents several 
advantages compared to the brute force, discretization-based, alternative.

Firstly, \4d\ is a fast and frugal algorithm; its good
performances and scalability have been successfully demonstrated on real-world
problems and on large-sized artificial datasets \cite{IJCAI05}.
Secondly, data mining applications specifically involve two key 
steps, exemplified in this paper: i) 
understanding the expert's goals and requirements; ii) tuning the 
parameters involved in the specifications. With regard to both 
steps, the ability of Evolutionary Computation to 
work under bounded resources is a 
very significant advantage. Evolutionary algorithms intrinsically 
are any-time algorithms, allowing the user 
to check at a low cost whether the process can deliver useful results, 
and more generally enabling her to control the trade-off between the 
computational resources needed and the quality of the results.

A main perspective for further research is to equip \4d\ with learning 
abilities, facilitating the automatic acquisition of the constraints and 
modelling the expert's expectations. 
A first step would be to automatically adjust the thresholds involved in
the constraints, based on the expert's feedback. Ultimately, the goal is to
design a truly user-centered mining system, combining advanced 
interactive optimization \cite{Combating}, 
online learning \cite{Online} 
and visual data mining 
\cite{Keim}. 

\subsection*{Acknowledgments}
We heartily thank Sylvain Baillet, Cognitive Neurosciences and Brain Imaging 
Lab., La Pitié Salpétrière and CNRS, who
provided the data and the interpretation of the results.
The authors gratefully acknowledge the support of the Pascal Network of Excellence (IST 2002506778).

\bibliographystyle{abbrv}

{
\small 
\begin{thebibliography}{10}

\bibitem{Online}
N.~Cesa-Bianchi, A.~Conconi, and C.~Gentile.
\newblock On the generalization ability of on-line learning algorithms.
\newblock {\em IEEE Transactions on Information Theory}, 50(9):2050--2057,
  2004.


\bibitem{Padraic}
D.~Chudova, S.~Gaffney, E.~Mjolsness, and P.~Smyth.
\newblock Translation-invariant mixture models for curve clustering.
\newblock In {\em Proc. of the Ninth Int. Conf.
  on Knowledge Discovery and Data Mining}, pages 79--88. ACM, 2003.

\bibitem{Corne}
D.~Corne, J.~D. Knowles, and M.~J. Oates.
\newblock The {P}areto envelope-based selection algorithm for multi-objective
  optimisation.
\newblock In {\em Proc. of PPSN - VI}, LNCS, pages 839--848. Springer Verlag, 2000.

\bibitem{DaidaGECCO99}
J.~Daida.
\newblock Challenges with verification, repeatability, and meaningful
  comparison in genetic programming: Gibson's magic.
\newblock In {\em Proc. of GECCO 99},
  pages 1069--1076. Morgan Kaufmann, 1999.

\bibitem{Debbook}
K.~Deb.
\newblock {\em Multi-Objective Optimization Using Evolutionary Algorithms}.
\newblock John Wiley, 2001.


\bibitem{Hastie}
T.~Hastie, R.~Tibshirani, and J.~H. Friedman.
\newblock {\em The Elements of Statistical Learning: Data Mining, Inference,
  and Prediction}.
\newblock Springer Series in Statistics, 2001.

\bibitem{ICA}
A.~Hyvarinen, J.~Karhunen, and E.~Oja.
\newblock {\em Independent Component Analysis}.
\newblock Wiley New York, 2001.

\bibitem{Hamalainen}
M.~Hämäläinen, R.~Hari, R.~Ilmoniemi, J.~Knuutila, and O.~V. Lounasmaa.
\newblock Magnetoencephalography: theory, instrumentation, and applications to
  noninvasive studies of the working human brain.
\newblock {\em Rev. Mod. Phys}, 65:413--497, 1993.

\bibitem{Keim}
D.~A. Keim, J.~Schneidewind, and M.~Sips.
\newblock Circleview: a new approach for visualizing time-related
  multidimensional data sets.
\newblock In {\em Proc. of Advanced Visual Interfaces}, pages 179--182. ACM
  Press, 2004.

\bibitem{BCIPhD}
T.~Lal.
\newblock {\em Machine Learning Methods for Brain-Computer Interfaces}.
\newblock PhD thesis, Max Plank Institute for Biological Cybernetics, 2005.

\bibitem{Laumans}
M.~Laumanns, L.~Thiele, K.~Deb, and E.~Zitsler.
\newblock Combining convergence and diversity in evolutionary multi-objective
  optimization.
\newblock {\em Evolutionary Computation}, 10(3):263--282, 2002.

\bibitem{Multi-modal}
J.-P. Li, M.~E. Balazs, G.~T. Parks, and P.~J. Clarkson.
\newblock A species conserving genetic algorithm for multimodal function
  optimization.
\newblock {\em Evolutionary Computation}, 10(3):207--234, 2002.

\bibitem{Combating}
X.~Llor{\`a}, K.~Sastry, D.~E. Goldberg, A.~Gupta, and L.~Lakshmi.
\newblock Combating user fatigue in IGAs: partial ordering, support vector
  machines, and synthetic fitness.
\newblock In {\em Proc. of GECCO 05}, pages
  1363--1370. ACM, 2005.

\bibitem{Bengio}
I.~McCowan, D.~Gatica-Perez, S.~Bengio, G.~Lathoud, M.~Barnard, and D.~Zhang.
\newblock Automatic analysis of multimodal group actions in meetings.
\newblock {\em IEEE Trans. on Pattern Analysis and Machine Intelligence
  (PAMI)}, 27(3):305--317, 2005.

\bibitem{Baillet}
D.~Pantazis, T.~E. Nichols, S.~Baillet, and R.~Leahy.
\newblock A comparison of random field theory and permutation methods for the
  statistical analysis of {MEG} data.
\newblock {\em Neuroimage}, 25:355--368, 2005.

\bibitem{TEMPORAL}
J.~Roddick and M.~Spiliopoulou.
\newblock A survey of temporal knowledge discovery paradigms and methods.
\newblock {\em IEEE Trans. on Knowledge and Data Engineering},
  14(4):750--767, 2002.

\bibitem{IJCAI05}
M.~Sebag, N.~Tarrisson, O.~Teytaud, S.~Baillet, and J.~Lefevre.
\newblock A multi-objective multi-modal optimization approach for mining stable
  spatio-temporal patterns.
\newblock In {\em Proc. of Int. Joint Conf. on AI, IJCAI'05}, pages
  859--864, 2005.

\bibitem{Spatial}
S.~Shekhar, P.~Zhang, Y.~Huang, and R.~R. Vatsavai.
\newblock Spatial data mining.
\newblock In H.~Kargupta and A.~Joshi, editors, {\em Data Mining: Next
  Generation Challenges and Future Directions}. AAAI/MIT Press, 2003.

\bibitem{Yu}
K.~Wu, S.~Chen, and P.~Yu.
\newblock Interval query indexing for efficient stream processing.
\newblock In {\em ACM Conf. on Information and Knowledge Management}, pages
  88--97. ACM Press, 2004.

\end{thebibliography}
}
\end{document}